# A Physics Enhanced Residual Learning (PERL) Framework for Vehicle Trajectory Prediction

Keke Long[1], Zihao Sheng[1], Haotian Shi[1,*], Xiaopeng Li[1,*], Sikai Chen[1], Sue Ahn[1]

[1] Department of Civil & Environmental Engineering, University of Wisconsin-Madison, Madison, Wisconsin, 53706, USA

Corresponding Authors: Haotian Shi, hshi84@wisc.edu; Xiaopeng Li, xli2485@wisc.edu

## ABSTRACT

While physics models for predicting system states can reveal fundamental insights due to their parsimonious structure, they may not always yield the most accurate predictions, particularly for complex systems. As an alternative, neural networks (NN) models usually yield more accurate predictions which however lack interpretable physical insights. To articulate the advantages of both physics and NN models while circumventing their limitations, this study proposes a Physics-Enhanced Residual Learning (PERL) framework that adjusts a physics model prediction with a corrective residual predicted from a residual learning NN model. The integration of the physics model preserves the interpretability and tremendously reduces the amount of training data compared to pure NN models. We apply PERL to a vehicle trajectory prediction problem with real-world trajectory data, where we use an adapted Newell car-following model as the physics model and the Long Short-Term Memory (LSTM) model as the residual learning model. We compare this PERL model with pure physics models, NN models, and other physics-informed neural network (PINN) models. The result reveals that PERL yields the best prediction when the training data is small. Besides, the PERL model has fast convergence during training. Moreover, the PERL model requires fewer parameters to achieve similar predictive performance compared to NN and PINN models. Sensitivity analysis shows the PERL model consistently outperforms other models using different physics and residual learning models.

# 1 INTRODUCTION

The prediction approach in recent decades has undergone a revolutionary change with the development of computer technology. Traditionally, researchers used theoretical derivation combined with experimental verification to study natural phenomena and then make predictions based on the theory they learned. Physics models are constructed from fundamental laws and principles governing physical systems for system state prediction (Figure 1A). As much as it intuitively reveals fundamental insights into the nature of the system, the parsimonious structures of a physics model may not always accurately describe complex nonlinearity, high dimensionality, or latent patterns, thus limiting its predictability.

The parsimony of physics models sometimes was also a compromise when only limited data and computational resources were available. The advent of data science leads to a prolific accumulation of data across various fields. Concurrently, neural network (NN) models along with numerous data-driven machine learning methods have emerged to enhance predictability, as it is shown that NN models are capable of approximating any continuous function (Figure 1B) [1]. NN models have been extensively utilized in domains such as medical forecasting [2], weather forecasting [3], and road traffic prediction [4,5], the third being the focused domain of this study.

Despite the superior predictability of NN models, they suffer from several limitations in real-world applications, including lack of interpretability, lack of robustness, and dependence on vast data [6,7]. To overcome these shortcomings, the physics-informed neural network (PINN) model (Figure 1C) has emerged [7,8]. To enhance the interpretability and ensure the stability of models beyond the domains of their training data, the PINN model encodes physics prior to NN. The widely adopted PINN for road traffic prediction, such as vehicle trajectory prediction, integrates physics priors into the NN's loss function [9–12]. This approach strikes a balance between the physics model and NN by modulating the weight of the deviation from the physics model prediction and that of NN's loss function. Despite salient advantages, the deviation correction towards the physics model prediction that is often less accurate may somehow compromise PINN's predictability. Another concern arises from unstable training due to different scales and convergence rates between the NN loss and the physics model deviation. Thus, tuning the weights of the physics model within the loss function is proven to be challenging [13–15]. Furthermore, without sufficient data, PINN may underperform since the efficacy of PINNs heavily leans toward the relatively inferior predictability of the physics model.

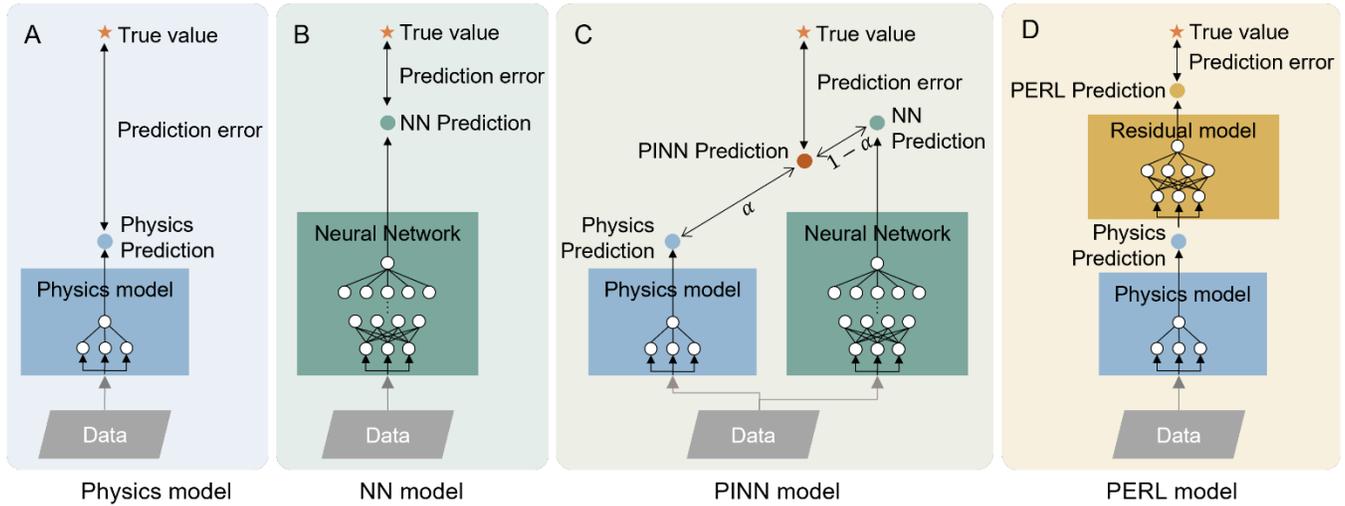

**Figure 1 General structure of PERL model and three existing models: Physics model, NN model, PINN model.**

To harvest the advantages of all these models while circumventing their limitations, we propose a novel Physics-Enhanced Residual Learning (PERL) framework (Figure 1D). Instead of a compromise between the two models, PERL further improves the physics model prediction with a residual term learned from the NN model. The physics model component alone can yield a reasonably accurate (albeit not the best) prediction with only a parsimonious structure specified by a few parameters. Then, as opposed to a pure NN model starting from scratch, the residual learning model component only needs to improve an already reasonably accurate prediction to a higher accuracy comparable to the pure NN model, which likely takes much less training data and computational resources and thus only needs fewer parameters and less training data. Meanwhile, unlike PINN which may draw the prediction toward relatively less accurate physics model prediction, PERL is pushing the prediction away from the physics model prediction to the most accurate value that a data-driven model can ever reach. With this, PERL better integrates the merits of both physics and NN models to achieve *the best and uncompromised predictability* and *the* *fastest convergence rate* with *the fewest parameters* and *inherited interpretability*. While the concept of a residual term has been explored in various fields, such as physics systems modeling [16,17], robotic manipulation tasks [18,19], and electromagnetic modeling [20], to the best of the authors' knowledge, residual learning has rarely been explored for system state prediction, which motivated us to develop the PERL model.

To demonstrate the effectiveness of the PERL model, we applied the PERL framework to the vehicle trajectory prediction problem using a real-world dataset. Vehicle trajectory prediction in the context of road traffic is a procedure that uses observed motions of road vehicles to predict their near-future trajectory attributes such as speed and acceleration. It is a necessary step for an automated vehicle to predict the motion paths of its surrounding vehicles and ensure safety control. Further, with the development of high-resolution traffic sensing technologies, vehicle trajectory prediction can be applied to predict future traffic states and potential traffic conflicts for safer, faster, and greener traffic operations. Road traffic dynamics is a representative example of complex systems with partially known physics prior. For example, some general physics rules that the following vehicle behavior is influenced by the behavior of the previous vehicle, are already considered in those widely recognized car-following models [21,22]. In this application, we first obtain a calibrated physics model based on the Next-Generation Simulation (NGSIM) dataset [23]. The residuals obtained from this physics model component are utilized to train a residual learning model component. The final prediction of the PERL model is the summation of the physics prediction and the learned residual. The performance of the PERL model is compared with the pure physics model, the pure NN model, and the PINN model. To further validate the generalization capability of the PERL framework, we conducted a sensitivity analysis to assess PERL's performance when paired with alternative residual learning models and physics car-following models.

## 2 METHODOLOGY

This section describes the prediction problem and the PERL model. Denote the input state (e.g., traffic states) space and the output response (e.g., set of longitudinal accelerations of the subject vehicle) space of a physics system by $\mathcal{S}$ and $\mathcal{Y}$, respectively, where $M$ and $N$ denote the numbers of dimensions of these two spaces, and $\mathbb{R}$ is the set of real numbers, respectively. Let $g(\cdot): \mathcal{S} \to \mathcal{Y}$ denote the ground-truth system that has response $g(s)$ for each input state $s \in \mathcal{S}$. The prediction problem is to construct a prediction function $f(\cdot|\theta): \mathcal{S} \to \mathcal{Y}$ such that $f(s|\theta)$ is the prediction of $g(s), \forall s \in \mathcal{S}$, where $\theta \in \Theta$ represents a tunable parameter vector in a parameter space $\Theta$. The predictability of function $f(\cdot|\theta)$ is evaluated by the prediction error between $f(s|\theta)$ and $g(s), \forall s \in \mathcal{S}$, e.g., the mean squared error (MSE). Solving a prediction problem is usually to find the optimal $\theta$ to minimize the prediction error or maximize its predictability. The interpretability of $f(\cdot|\theta)$ is characterized by the ability of this function to reflect comprehensible physics rules and relationships e.g., how elements in $\theta$ can be interpreted as certain physics rules and relationships.

## 2.1 PERL

Built upon the defined prediction problem, the PERL model can be described as follows. The PERL prediction is the combination of a physics model component and a residual learning component, formulated as $f^{\text{PERL}}(\cdot|\theta^{\text{PERL}}) := f^{\text{Phy}}(\mathcal{S}^{\text{Phy}}(\cdot)|\theta^{\text{Phy}}) + f^{\text{RL}}(\cdot|\theta^{\text{RL}})$. The components of this formulation are specified as follows. $f^{\text{Phy}}(\mathcal{S}^{\text{Phy}}(s)|\theta^{\text{Phy}}): \mathcal{S}^{\text{Phy}} \to \mathcal{Y}^{\text{Phy}}$ is the physics model component that makes an initial prediction of an input state $s \in \mathcal{S}$. Here, $\mathcal{S}^{\text{Phy}}(s) \in \mathcal{S}^{\text{Phy}}$ denote the projection of $s \in \mathcal{S}$ to $\mathcal{S}^{\text{Phy}}$ as a subspace of $\mathcal{S}$. Note that due to parsimony, the input state to the physics model component $f^{\text{Phy}}$ may not include all dimensions in $\mathcal{S}$ and can be a projection to its subspace $\mathcal{S}^{\text{Phy}}$. $\theta^{\text{Phy}}$ is the parameter vector for $f^{\text{Phy}}$ in parameter space $\Theta^{\text{Phy}}$, which is often a low-dimensional space due to the parsimony of the physics model. $\mathcal{Y}^{\text{Phy}}$ is the range of component $f^{\text{Phy}}$ with domain $\mathcal{S}^{\text{Phy}}$ and shall have the same cardinality as $\mathcal{Y}$ since $f^{\text{Phy}}$ also predicts $g$. $f^{\text{RL}}(\cdot|\theta^{\text{RL}}): \mathcal{S} \to \mathcal{Y}^{\text{RL}} := \mathcal{Y} - \mathcal{Y}^{\text{Phy}}$ is the residual learning component such that $f^{\text{RL}}(s|\theta^{\text{RL}})$ predicts the residual from the physical model prediction to the ground truth, i.e., $g(s) - f^{\text{Phy}}(\mathcal{S}^{\text{Phy}}(s)|\theta^{\text{Phy}}), \forall s \in \mathcal{S}$. $\theta^{\text{RL}}$ is the parameter vector for the residual learning component $f^{\text{RL}}$ in parameter space $\Theta^{\text{RL}}$. With this, the PERL model prediction $f^{\text{PERL}}(s|\theta^{\text{PERL}}) \in \mathcal{Y}$ is the summation of the physics model prediction $f^{\text{Phy}}(\mathcal{S}^{\text{Phy}}(s)|\theta^{\text{Phy}})$ and the residual prediction $f^{\text{RL}}(s|\theta^{\text{RL}}), \forall s \in \mathcal{S}$, where the parameter vector $\theta^{\text{PERL}}$ is a simple concatenation of those for the physics and residual learning components, i.e., $\theta^{\text{PERL}} := [\theta^{\text{Phy}}, \theta^{\text{RL}}] \in \Theta^{\text{Phy}} \times \Theta^{\text{RL}}$.

In a real-world application, let $\mathcal{J} := \{1, \cdots, I\}$ denote the index set of observed samples, where $I \in \mathbb{R}^+$ is the total number of samples. $\mathcal{J}$ is randomly divided into three sets, i.e., $\mathcal{J}^{\text{Train}} \cup \mathcal{J}^{\text{Val}} \cup \mathcal{J}^{\text{Test}} = \mathcal{J}$, $\mathcal{J}^{\text{Train}} \cap \mathcal{J}^{\text{Val}} = \emptyset$; $\mathcal{J}^{\text{Train}} \cap \mathcal{J}^{\text{Test}} = \emptyset$; $\mathcal{J}^{\text{Val}} \cap \mathcal{J}^{\text{Test}} = \emptyset$. Define $\omega^{\text{Train}}$ and $\omega^{\text{Val}}$ as the proportion of the total dataset allocated for training and validation purposes: $|\mathcal{J}^{\text{Train}}| = \lfloor \omega^{\text{Train}} \times \mathcal{J} \rfloor$ and $|\mathcal{J}^{\text{Val}}| = \lfloor \omega^{\text{Val}} \times \mathcal{J} \rfloor$ where $\lfloor \cdot \rfloor$ denotes the floor function. The testing set index $\mathcal{J}^{\text{Test}}$ consists of the remaining indexes, calculated as $|\mathcal{J}^{\text{Test}}| = |\mathcal{J}| - |\mathcal{J}^{\text{Train}}| - |\mathcal{J}^{\text{Val}}|$. $\theta^{\text{Phy}}$ is obtained by calibration using a set of observed states $\mathcal{S}^{\text{Train}} := \{s_i \in \mathcal{S}\}_{i \in \mathcal{J}^{\text{Train}}}$: $\theta^{*\text{Phy}} := \underset{\theta^{\text{Phy}} \in \Theta^{\text{Phy}}}{\operatorname{argmin}} \sum_{s \in \mathcal{S}^{\text{Train}}} \left( f^{\text{Phy}}(S^{\text{Phy}}(s) | \theta^{\text{Phy}}) - g(s) \right)^2$. $\theta^{\text{RL}}$ is obtained by training with training dataset: $\theta^{*\text{RL}} = \underset{\theta^{\text{RL}} \in \Theta^{\text{RL}}}{\operatorname{argmin}} \sum_{s \in \mathcal{S}^{\text{Train}}} \left( f^{\text{RL}}(s | \theta^{\text{RL}}) - \left( g(s) - f^{\text{Phy}}(S^{\text{Phy}}(s) | \theta^{*\text{Phy}}) \right) \right)^2$.

**2.2 Baseline models**

For comparison purposes, we also consider a pure NN prediction model, a pure physics prediction model, and a PINN model.

*2.2.1 Physics model*

A physics model (e.g., a car following model) is denoted as $f^{\text{Phy}}(S^{\text{Phy}}(s) | \theta^{\text{Phy}}) : \mathcal{S}^{\text{Phy}} \to \mathcal{Y}^{\text{Phy}}$. Here $S^{\text{Phy}}(s) \in \mathcal{S}^{\text{Phy}}$ denote the projection of $s \in \mathcal{S}$ to $\mathcal{S}^{\text{Phy}}$ as a subspace of $\mathcal{S}$. $\mathcal{Y}^{\text{Phy}}$ is the range of component $f^{\text{Phy}}$ with domain $\mathcal{S}^{\text{Phy}}$ and have the same cardinality as $\mathcal{Y}$. $\theta^{\text{Phy}}$ is the parameter vector for $f^{\text{Phy}}$ in a low-dimensional parameter space $\Theta^{\text{Phy}}$ and usually obtained by calibration:

$$\theta^{*\text{Phy}} = \underset{\theta^{\text{Phy}} \in \Theta^{\text{Phy}}}{\operatorname{argmin}} \sum_{s \in \mathcal{S}^{\text{Train}}} \left( f^{\text{Phy}}(S^{\text{Phy}}(s) | \theta^{\text{Phy}}) - g(s) \right)^2 \qquad (1)$$

*2.2.2 NN model*

An NN model is denoted as $f^{\text{NN}}(s | \theta^{\text{NN}}) : \mathcal{S} \to \mathcal{Y}^{\text{NN}}$ that makes a prediction of an input state $s \in \mathcal{S}$. $\mathcal{Y}^{\text{NN}}$ is the range of component $f^{\text{NN}}$ with domain $\mathcal{S}$ and have the same cardinality as $\mathcal{Y}$. $\theta^{\text{NN}}$ is the parameter vector for $f^{\text{NN}}$ in parameter space $\Theta^{\text{NN}}$ and obtained by training using the training dataset:

$$\theta^{*\text{NN}} = \underset{\theta^{\text{NN}} \in \Theta^{\text{NN}}}{\operatorname{argmin}} \sum_{s \in \mathcal{S}^{\text{Train}}} \left( f^{\text{NN}}(s | \theta^{\text{NN}}) - g(s) \right)^2 \qquad (2)$$

*2.2.3 PINN model*

A PINN model denoted as $f^{\text{PINN}}(s|\theta^{\text{PINN}}): \mathcal{S} \to \mathcal{Y}^{\text{PINN}}$ that makes a prediction of an input state $s \in \mathcal{S}$. $\mathcal{Y}^{\text{PINN}}$ is the range of component $f^{\text{PINN}}$ with domain $\mathcal{S}$ and have the same cardinality as $\mathcal{Y}$. The PINN model adopted the physics model component $f^{\text{Phy}}(S^{\text{Phy}}(s)|\theta^{\text{Phy}})$ to help the training process of its NN component $f^{\text{NN}}(s|\theta^{\text{NN}})$. The physics model cooperated with the loss function of the NN component during training, where the NN model prediction should be close to both the physics model prediction and the ground truth. Thus, the loss function of the NN component contains two parts $\text{LOSS}_1$ and $\text{LOSS}_2$:

$$\text{LOSS}_1 \coloneqq \left(f^{\text{NN}}(s|\theta^{\text{NN}}) - g(s)\right)^2 \qquad (3)$$

$$\text{LOSS}_2 \coloneqq \left(f^{\text{NN}}(s|\theta^{\text{NN}}) - f^{\text{Phy}}(S^{\text{Phy}}(s)|\theta^{\text{Phy}})\right)^2 \qquad (4)$$

$$\theta^{\text{PINN}} = \underset{\theta^{\text{PINN}} \in \Theta^{\text{PINN}}}{\text{argmin}} \sum_{s \in \mathcal{S}^{\text{Train}}} \mu \cdot \text{LOSS}_1 + (1-\mu) \cdot \text{LOSS}_2 \qquad (5)$$

where $\mu$ is the weight hyperparameters to be tuned. The parameter vector $\theta^{\text{PINN}} \in \Theta^{\text{PINN}}$ is a concatenation of the physics component and NN component, i.e., $\theta^{\text{PINN}} \coloneqq [\theta^{\text{Phy}}, \theta^{\text{NN}}] \in \Theta^{\text{Phy}} \times \Theta^{\text{NN}}$.

## 3  USE CASE: VEHICLE TRAJECTORY PREDICTION.

We applied the PERL model to the vehicle trajectory prediction problem as described below. As shown in Figure 2, an input state stems from a set of $K$ consecutive vehicle trajectories, indexed by $k \in \mathcal{K} \coloneqq \{1,2,\cdots,K\}$ from the upstream lead vehicle indexed by 1 through the downstream ego vehicle by $K$, at $T^{\text{b}}$ historical time points $\mathcal{T}^{\text{b}} \coloneqq \{t_0 - (T^{\text{b}}-1)\delta, \cdots, t_0 - \delta, t_0\}$ separated by time interval $\delta$. The input state is defined as $s = [a_{kt}, v_{kt}, \Delta d_{kt}]_{\forall k \in \mathcal{K}, t \in \mathcal{T}^{\text{b}}}$ where $a_{kt}, v_{kt}$, and $\Delta d_{kt}$ are the acceleration, speed, and spacing (from the preceding vehicle) of vehicle $k$ at time $t$ in sample $i$, respectively. With this, $|\mathcal{S}| = 3|\mathcal{K}|T^{\text{b}}$. The corresponding output response $g(s)$ is the ground-truth trajectory of ego vehicle $K$ at future time points $\mathcal{T}^{\text{f}} \coloneqq \{t_0 + \delta, t_0 + 2\delta, \cdots, t_0 + T^{\text{f}}\delta\}$, i.e., $g(s) = [a_{Kt}]_{t \in \mathcal{T}^{\text{f}}}$ where $a_{Kt}$ is the ground-truth acceleration of ego vehicle $K$ at a future time point $t$. With this, $|\mathcal{Y}| = T^{\text{f}}$.

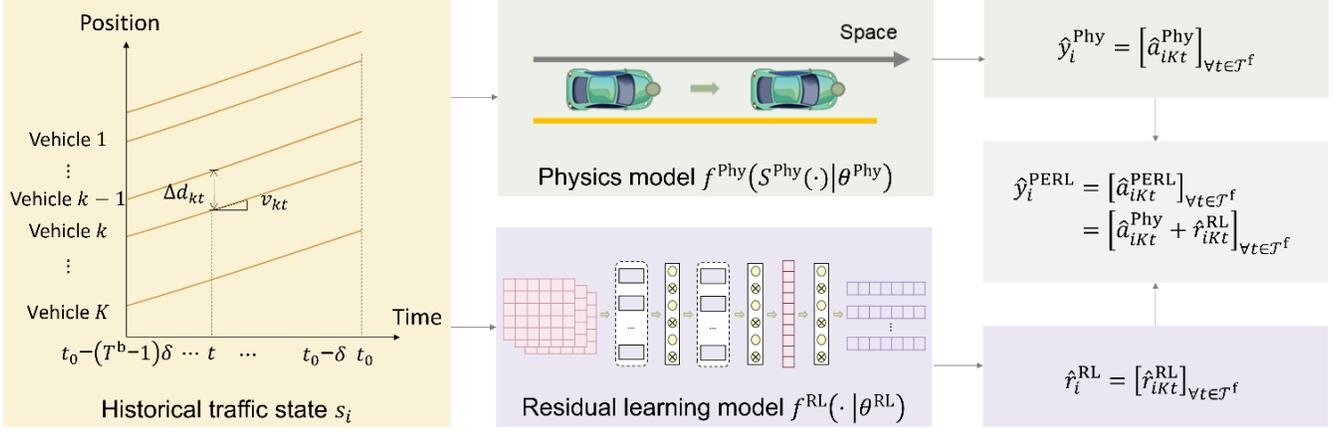

**Figure 2 Workflow of the PERL model (example in vehicle trajectory prediction).**

In the physics component, $S^{\text{Phy}}(\cdot)$ projects $s$ to a much lower-dimensional space, i.e., $S^{\text{Phy}}(s) = [a_{kt_k}, v_{kt_k}, \Delta d_{kt_k}]_{k \in \mathcal{K}, t_k \in [t_0 - \tau_k, \cdots, t_0]}$ or its subset where $\tau_k$ is a short reaction time depending on the specific physics model $t_0 - \tau_k \in \mathcal{T}^{\text{b}}$. With this, $|\Theta^{\text{Phy}}|$ is on the order of $|S^{\text{Phy}}(s)|$, i.e., $O(|S^{\text{Phy}}(s)|)$, and thus much smaller than $|\mathcal{S}|$. $\mathcal{Y}^{\text{Phy}}$ is the predicted acceleration of vehicle $K$: $\left[\hat{a}_{Kt}^{\text{Phy}}(\theta^{\text{Phy}})\right]_{\forall t \in \mathcal{T}^{\text{f}}} = f^{\text{Phy}}(S^{\text{Phy}}(s)|\theta^{\text{Phy}})$. therefore, we get the residual of acceleration predicted by the physics model $r_{Kt}(\theta^{\text{Phy}}) \coloneqq g(s) - \hat{a}_{Kt}^{\text{Phy}}(\theta^{\text{Phy}}), \forall t \in \mathcal{T}^{\text{f}}$.

In the residual learning component, $|\Theta^{\text{RL}}|$ is generally polynomial in relation to $|s|$ and the architecture of the model, thus $|\Theta^{\text{RL}}|$ is likely much greater than $|\Theta^{\text{Phy}}|$. $\mathcal{Y}^{\text{RL}}$ is the predicted residual of vehicle $K$: $r_{Kt}$, i.e., $\left[\hat{r}_{Kt}^{\text{RL}}(\theta^{\text{RL}})\right]_{\forall t \in \mathcal{T}^{\text{f}}} = f^{\text{RL}}(s|\theta^{\text{RL}})$. Therefore, the output of the PERL model is the $f^{\text{PERL}}(s|\theta^{\text{PERL}}) = \left[\hat{a}_{Kt}^{\text{Phy}}(\theta^{\text{Phy}}) + \hat{r}_{Kt}^{\text{RL}}(\theta^{\text{RL}})\right]_{\forall t \in \mathcal{T}^{\text{f}}}$.

## 4 NUMERICAL EXAMPLE

We adopt the NGSIM US101 data[23] for a numerical validation of the PERL model. A total of $I = 20,000$ samples, $\delta = 0.1s$, $\omega^{\text{Train}} = 0.6$, $\omega^{\text{Val}} = 0.2$. Each sample comprise $K = 4$ consecutive vehicles. For one-step prediction: $T^{\text{b}} = 50$, $T^{\text{f}} = 1$; for multi-step prediction, $T^{\text{b}} = T^{\text{f}} = 50$. The predicted speed of vehicle $K$ at time $t$ in sample $i \in \mathcal{J}$ is calculated as $\hat{v}_{iKt} = v_{iKt_0} + \sum_{t'=t_0}^{t} \hat{a}_{iKt'}\delta, \forall t' \in \mathcal{T}^{\text{f}}$. The evaluation for one-step and multi-step prediction on the test set $\mathcal{J}^{\text{test}}$ includes two metrics: the MSE of acceleration and speed prediction: $\text{MSE}_{\text{test}}^{a} = \frac{1}{|\mathcal{J}^{\text{Test}}|T^{\text{f}}} \sum_{i \in \mathcal{J}^{\text{Test}}, t \in \mathcal{T}^{\text{f}}} (a_{iKt} - \hat{a}_{iKt})^2$ and $\text{MSE}_{\text{test}}^{v} = \frac{1}{|\mathcal{J}^{\text{Test}}|T^{\text{f}}} \sum_{i \in \mathcal{J}^{\text{Test}}, t \in \mathcal{T}^{\text{f}}} (v_{iKt} - \hat{v}_{iKt})^2$. The convergence of training process is measured by the change of the MSE of acceleration and speed prediction on the validation set: $\text{MSE}_{\text{Val}}^{a} = \frac{1}{|\mathcal{J}^{\text{Val}}|T^{\text{f}}} \sum_{i \in \mathcal{J}^{\text{Val}}, t \in \mathcal{T}^{\text{f}}} (a_{iKt} - \hat{a}_{iKt})^2$ and $\text{MSE}_{\text{Val}}^{v} = \frac{1}{|\mathcal{J}^{\text{Val}}|T^{\text{f}}} \sum_{i \in \mathcal{J}^{\text{Val}}, t \in \mathcal{T}^{\text{f}}} (v_{iKt} - \hat{v}_{iKt})^2$.

### 4.1 PERL physics component

For the PERL method, a model adapted from Newell's car-following model[35] is adopted as the physics model, and the Long Short-Term Memory (LSTM) model[36,37] as the residual learning model.

For the physical models, we employed three different approaches in this study. Initially, we utilized an adapted Newell model which is characterized by a single parameter $w$ denoting the wave speed,[38] shown in Figure 3. The physics model output $\mathcal{Y}^{\text{Phy}} = \left[\hat{a}_{iKt}^{\text{Phy}}(\theta^{\text{Phy}})\right]_{\forall t \in \mathcal{T}^{\text{f}}, i \in \mathcal{J}}$ is given by:

$$\hat{a}_{iKt}^{\text{Phy}} = a_{ik't - \frac{D_{iKk't}}{w}}, \forall k' \in \{1, 2, \cdots, K-1\}, t \in \mathcal{T}^{\text{f}}, i \in \mathcal{J} \tag{6}$$

where $D_{iKk't}$ represents the position distance between vehicle $K$ and vehicle $k'$ at time $t \in \mathcal{T}^{\text{f}}$ in sample $i \in \mathcal{J}$.

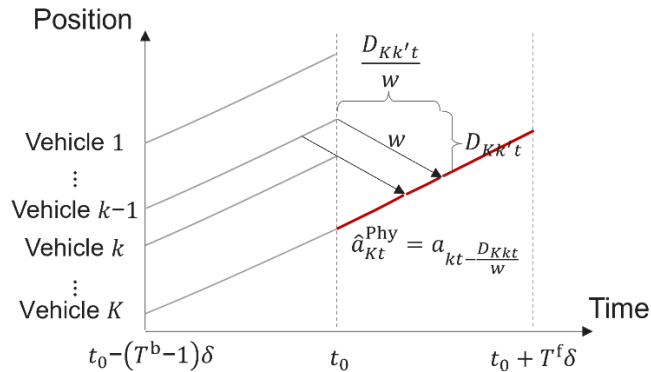

**Figure 3 Adapted Newell model for vehicle trajectory prediction.**

For sensitivity analysis, we also adopt the IDM model and FVD model as an alternative physics model. IDM model provides a model acceleration function as a continuous function of speed, gap, and speed difference and is expressed as follows:

$$\hat{a}_{iK(t_0+1)}^{Phy} = \bar{a}\left[1 - \left(\frac{v_{iKt_0}}{v^f}\right)^4 - \left(\frac{S(v_{iKt_0}, \Delta v_{iKt_0})}{\Delta x_{Kt_0}}\right)^2\right], \forall i \in \mathcal{I} \quad (7)$$

$$S(v_{iKt_0}, \Delta v_{iKt_0}) = S_0 + T^g v_{iKt_0} - \frac{v_{iKt_0} \cdot \Delta v_{iKt_0}}{2\sqrt{\bar{a}\bar{b}}}, \forall i \in \mathcal{I} \quad (8)$$

where $\Delta d_{i,n,t}$ is the relative distance between two adjacent vehicles, $S(v_{iKt_0}, \Delta v_{iKt_0})$ is the desired space headway function and is calculated from the speed $v_{n,t}$ and the relative speed $\Delta v_{iKt_0} := v_{iKt_0} - v_{i(K-1)t_0}, \forall i \in \mathcal{I}$, $v^f$ is the free flow speed, $\bar{a}$ is the maximum acceleration, $\bar{b}$ is the comfortable deceleration, $S_0$ is the minimum space. $T^g$ is the desired time gap.

The FVD model focuses on the speed difference between vehicles, considering both relative speeds and spatial gaps. This model provides a comprehensive representation of vehicle dynamics, especially in high-density traffic scenarios where speed differences play a pivotal role.

$$\hat{a}_{iK(t_0+1)}^{Phy} = \kappa[V(\Delta d_{iKt_0}) - v_{iKt_0}] + \lambda \Delta v_{iKt_0}, \forall i \in \mathcal{I} \quad (9)$$

$$V(\Delta d_{iKt_0}) := V_1 + V_2 \tanh[C_1(\Delta d_{iKt_0} - l_c) - C_2], \forall i \in \mathcal{I} \quad (10)$$

where $\kappa$ and $\lambda$ are sensitivity parameters and $V(\Delta d_{iKt_0})$ is the optimal speed that the drivers prefer. We adopted the calibrated parameters for $V(\Delta d_{iKt_0})$ as $V_1 = 6.75 m/s, V_2 = 7.91 m/s, C_1 = 0.13 m^{-1}$, and $C_2 = 1.54$.

We employed the Monte Carlo method for calibration, which randomly selected a certain number of samples from the dataset for calibration. The calibrated parameters of physics models are shown in TABLE 1. As the training data size increased, the calibration results tended to stabilize and the calibrated parameter is near the value in the literature reference.[39,40] The calibrated parameters of the physics model will be used as initial values in the PINN model.

**TABLE 1 Calibrated parameters of physics models with different training sample sizes.**

| Model | Parameters | Training sample size | | | | | | |
|---|---|---|---|---|---|---|---|---|
| | | 300 | 500 | 1000 | 2000 | 5000 | 10000 | 12000 |
| Adapted Newell model | $w$ (m/s) | 4.044 | 4.15 | 4.122 | 4.027 | 4.03 | 4.01 | 4.01 |
| | (variance) | 0.088 | 0.017 | 0.016 | 0.018 | 0.001 | 0 | 0 |
| IDM | $v^f$ (m/s) | 22.821 | 22.512 | 22.084 | 22.377 | 22.125 | 22.125 | 22.495 |
| | (variance) | 0.035 | 0.095 | 0.013 | 0.129 | 0.003 | 0.003 | 0.217 |
| | $\bar{a}$ (m/s²) | 0.682 | 0.475 | 0.626 | 0.54 | 0.613 | 0.613 | 0.911 |
| | (variance) | 0.095 | 0.09 | 0.175 | 0.099 | 0.192 | 0.192 | 0 |
| | $\bar{b}$ (m/s²) | 2.562 | 2.271 | 2.473 | 2.262 | 2.478 | 2.478 | 2.859 |
| | (variance) | 0.205 | 0.141 | 0.246 | 0.087 | 0.237 | 0.237 | 0.003 |
| | $S_0$ (m) | 1.712 | 1.703 | 1.627 | 1.608 | 1.615 | 1.615 | 1.627 |
| | (variance) | 0.01 | 0.021 | 0.013 | 0.006 | 0.01 | 0.01 | 0.013 |
| | $T^g$ (s) | 1.2 | 1.216 | 1.23 | 1.162 | 1.266 | 1.266 | 1.132 |
| | (variance) | 0.005 | 0.002 | 0 | 0.02 | 0.003 | 0.003 | 0.017 |
| FVD | $k$ | 0.108 | 0.073 | 0.088 | 0.044 | 0.108 | 0.076 | 0.007 |
| | (variance) | 0.003 | 0.003 | 0.002 | 0.003 | 0.005 | 0.006 | 0 |
| | $\lambda$ | 0.498 | 0.397 | 0.478 | 0.452 | 0.445 | 0.284 | 0.137 |
| | (variance) | 0.019 | 0.041 | 0.041 | 0.076 | 0.029 | 0.037 | 0 |
| | $V_0$ (m/s) | 19.779 | 18.814 | 19.647 | 23.034 | 21.542 | 22.093 | 24.158 |
| | (variance) | 15.982 | 10.059 | 19.536 | 26.672 | 17.456 | 14.257 | 5 |
| | $b$ (m/s²) | 10.084 | 10.746 | 9.637 | 8.406 | 10.327 | 11.243 | 7.954 |
| | (variance) | 4.935 | 14.088 | 5.636 | 10.33 | 8.447 | 11.241 | 4.34 |
| | $\beta$ | 2.561 | 3.011 | 2.824 | 2.449 | 2.383 | 2.133 | 1.724 |
| | (variance) | 1.338 | 1.6 | 1.585 | 0.733 | 2.388 | 3.52 | 0.079 |

## 4.2 PERL residual learning component

For the NN model, we adopt the commonly used LSTM model as an example.[36,41] The LSTM model's input consists of $T^b$ time steps of acceleration, the speed, and the spacing (from the preceding vehicle) of $K$ consecutive vehicles. The model predicts $T^f$ steps of the subject vehicle's future acceleration. The LSTM model structure is shown in Figure 4. The NN model utilizes LSTM, consisting of an input layer, two LSTM layers with $\eta_1$ and $\eta_2$ units respectively, three Dropout layers with a dropout rate of 0.2 to prevent overfitting, a Dense layer (fully connected layer) with $\eta_3$ units, and a Rectified Linear Unit (ReLU) activation function that matches the dimensionality of the target output.

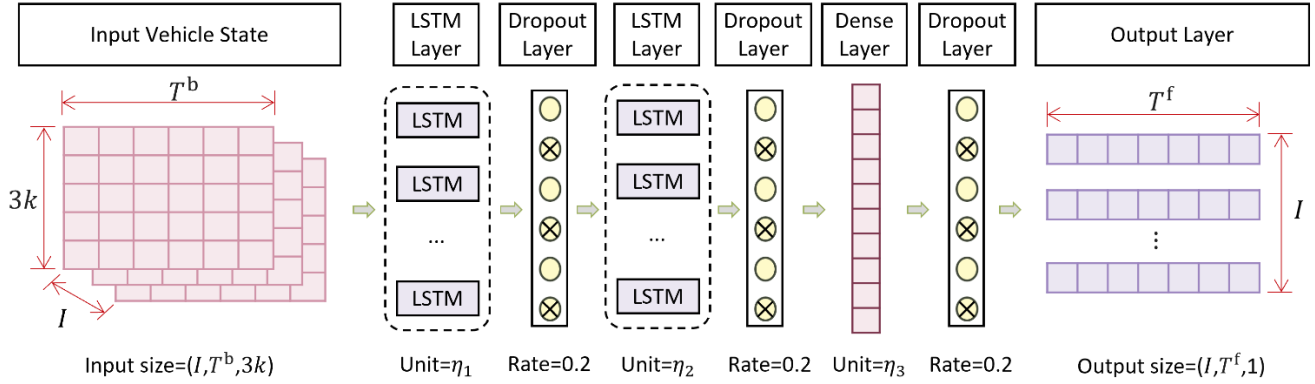

**Figure 4 LSTM-based NN model for vehicle trajectory prediction.**

In the sensitivity analysis, we also explored the use of the Gated Recurrent Unit (GRU) model.[42] GRU models are a kind of recurrent NN, which allow connections between neurons to form a directed cycle thus making it possible to exhibit dynamic temporal behavior. GRU has also been applied in car-following behavior modeling.[43] The designed GRU-based NN model has a structure closely resembling the LSTM-based NN model, shown in Figure 5, consisting of an input layer, two GRU layers with $\eta_1$ and $\eta_2$ units respectively, three Dropout layers with a dropout rate of 0.2, a Dense layer with $\eta_3$ units, and a ReLU activation layer.

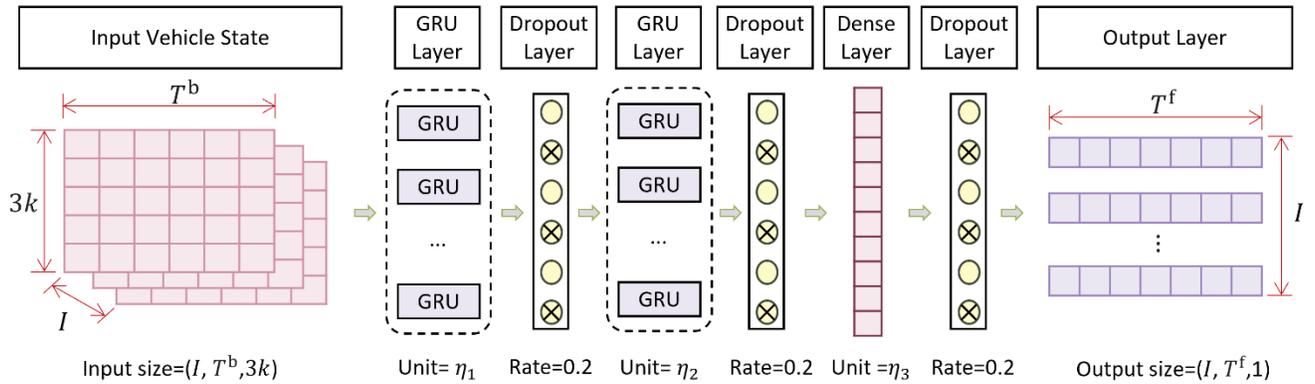

**Figure 5 GRU-based NN model for vehicle trajectory prediction.**

**4.3  Baseline models**

Regarding the three baseline models, the physics model employs the adapted Newell model [35], aligning with the physics component of PERL. The NN model utilizes the LSTM model, aligning with the residual component of PERL but with more parameters. The PINN model combines the adapted Newell model for its physics component and mirrors the LSTM model in its NN component.

**5  RESULTS**

**5.1  One-step comparison**

The results show that the PERL model exhibits superior predictability with less training data. In scenarios characterized by very limited data availability (e.g., 300 training data size) (Figure 6A 7B), both the PERL and PINN models outperform the NN model in terms of predictability. As the dataset expands beyond 1000 training samples, the performance differential between the PERL model and the baseline models, including the NN model, becomes markedly less pronounced (as depicted in Figure 6A and 7B). In this context, all models demonstrate remarkable predictability, achieving acceleration prediction errors as low as $\text{MSE}_{\text{Test}}^a = 0.047 \ m^2/s^4$.

**5.2  Multi-step prediction comparison**

In multi-step prediction, all models tend to show a diminished performance relative to one-step prediction, a trend that can be attributed to the heightened complexity inherent in predictions over longer time horizons. However, PERL still outperforms both PINN and NN models in scenarios with limited samples and provides superior velocity predictability compared to physics models (Figure 6C 7D).

One contributing factor to PERL's superior performance in predicting acceleration residuals, as opposed to the NN model's prediction of acceleration, is the different distribution patterns of acceleration residuals and acceleration. The variance in acceleration is greater than that of acceleration residuals across all samples, as depicted in Figure 6I. This difference is further highlighted by comparing the distributions of acceleration and acceleration residuals in a single sample, illustrated in Figure 6J and 7K. The observed lesser variance in acceleration residual suggests that in scenarios with limited data, the distribution of acceleration residual may more accurately mirror the actual distribution, facilitating better prediction outcomes. On the other hand, the higher variance in acceleration might lead to a less accurate representation of the true distribution in small sample scenarios, thereby potentially diminishing the prediction accuracy. These findings align with the operational principles of the PERL model, which chiefly engages in predicting the residual of the physics model, a task seemingly well-suited for situations with restricted data availability. This stands in contrast to the PINN model, which aims at the prediction of acceleration, a process that could be adversely affected by the noted higher variance, especially in data-scarce conditions. Hence, this visual representation further underscores the sample efficiency and effectiveness of the PERL model in acceleration prediction, particularly when data resources are limited.

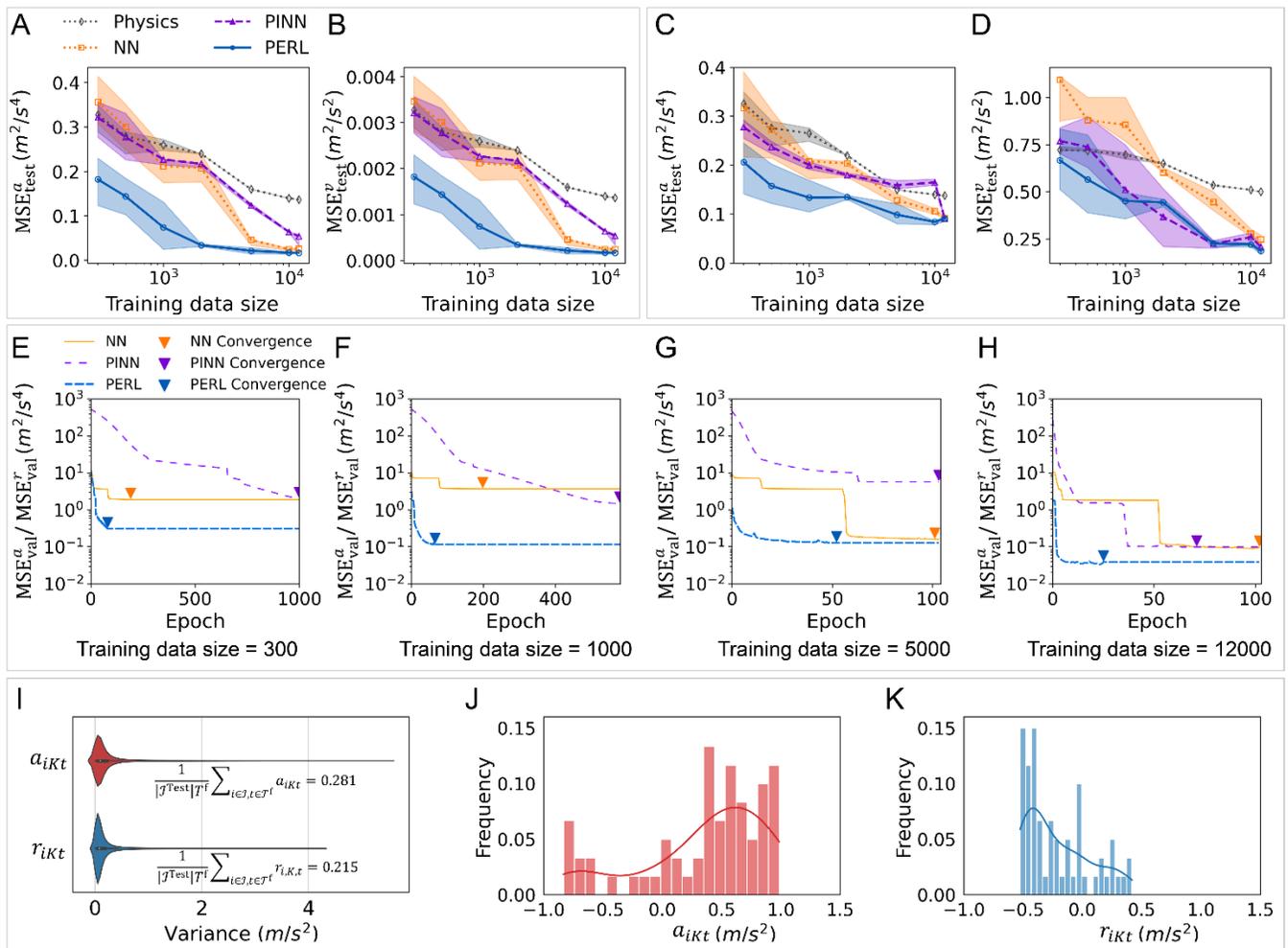

**Figure 6 PERL (LSTM+Adapted Newell) and baseline prediction performance**

Figure 6 shows when the training data size is 12000, the prediction accuracy of the PERL, PINN, and NN models was comparable, and all were superior to the physics model. This is validated by the 6 examples in Figure 7. Results indicate that both PINN and NN models can predict the overall trend of acceleration changes, but they fail to fit the local acceleration oscillation. The physics model can account for local acceleration oscillation based on physical principles, but as the physics model is calibrated using multiple vehicle trajectories, its results deviate significantly from real-world values. Only the PERL model is capable of not only predicting the overall trend of acceleration change but also keeping the local acceleration oscillation results from its physics model component.

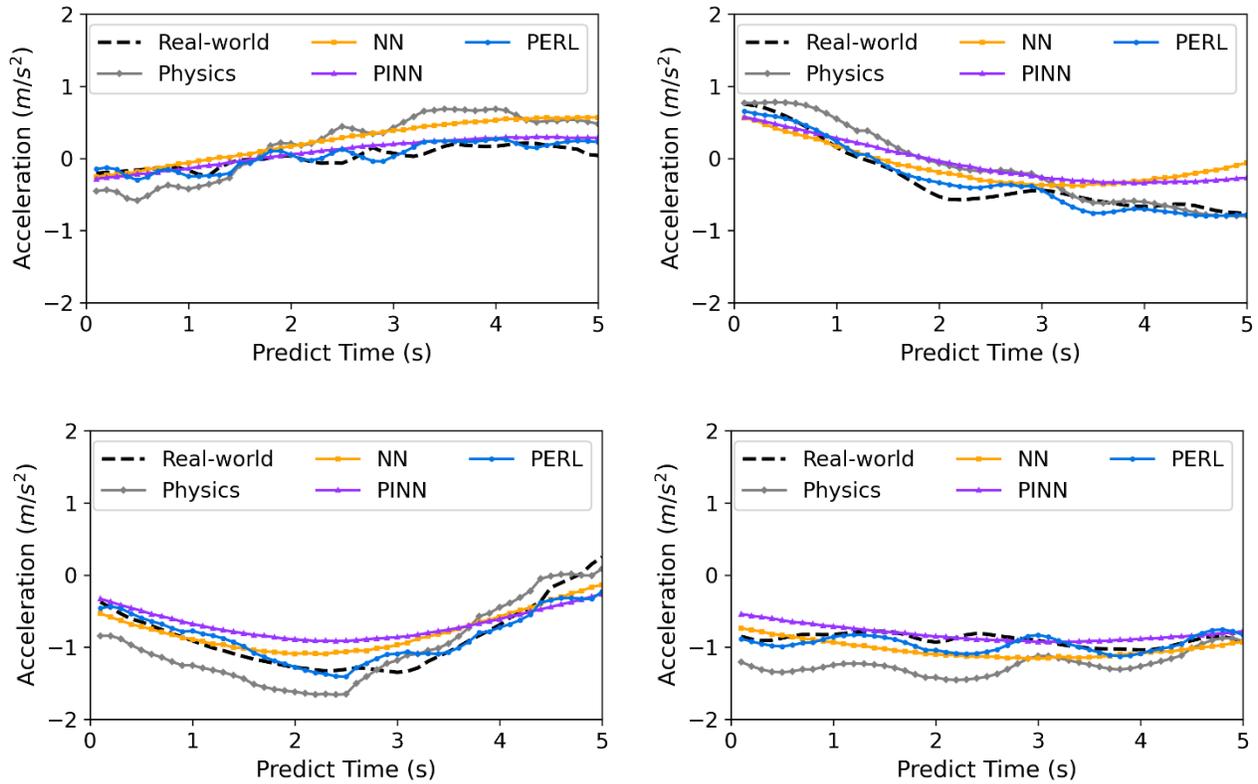

**Figure 7 Comparison of real-word acceleration and predicted result of PERL model and baseline models.**

## 5.3 Convergence comparison

To ensure a fair comparison of convergence among different models, we aligned the number of parameters in the NN, PINN, and PERL models. Specifically, we standardized the LSTM architecture within the NN component of the NN and PINN models and the residual learning component of the PERL model, setting both LSTM layers to have 128 and 64 units, respectively. The result shows that PERL converges more rapidly than both the NN and PINN models with various training data sizes (Figure 6E 7F 7G 7H). In scenarios with limited data (300 training data size), the PINN model converges at approximately 80 epochs, and the NN model converges at approximately 140 epochs, whereas the PINN model requires a substantially shorter time, around 1000 epochs (Figure 6E). This unstable training behavior of PINN arises because different parts of the loss function dominate the decrease at various times: sometimes it's influenced by the neural network and other times by the physics rules, this unbalanced back-propagated gradient results in an unstable learning process.[14] The NN model also converges slower than PERL despite having a near-identical number of parameters (Figure 6H). This is because the NN model needs to learn all the kinematic rules, while PERL only has to learn the features of the residuals. The convergence speed of PINN is also slower than that of the NN model because its objective function includes terms from both the neural network and the physics model, making the gradient search process more challenging.

This result highlights the advantage of the PERL model, which employs the same LSTM architecture and nearly identical number of parameters as the NN and PINN models but uses a more focused prediction approach that enables faster convergence and more efficient computation using fewer parameters.

## 5.4 Sensitivity analysis for PERL components

### 5.4.1 Physics car-following models

In single-step prediction tasks, a relatively straightforward task, it's feasible to employ car-following models that only consider the information of the preceding vehicle, such as the Intelligent Driver Model (IDM)[21] and the Full Velocity Difference (FVD) model[44]. We integrated the PERL model with the IDM and FVD models, utilizing them as the physics model, and paired them with an LSTM for residual learning.

PERL (LSTM+IDM) outperforms the physics model, the NN model (LSTM), and the PINN (LSTM+IDM) when the training data size is smaller than 10000 (Figure 8A 9B). This suggests the versatility of the PERL model, highlighting its compatibility with the IDM physics model. The IDM model effectively captures certain physics laws, and the residuals in acceleration prediction from the IDM model can be proficiently learned by the LSTM. The convergence of the IDM-based PERL (LSTM+IDM) is faster than both NN (LSTM) and PINN (LSTM+IDM) models (Figure 8C 9D). This emphasizes the capability of the IDM model to delineate a significant portion of the physics laws. Moreover, the IDM model can learn the acceleration residuals with less data, and its training convergence is faster than directly learning the acceleration using the NN model and PINN model.

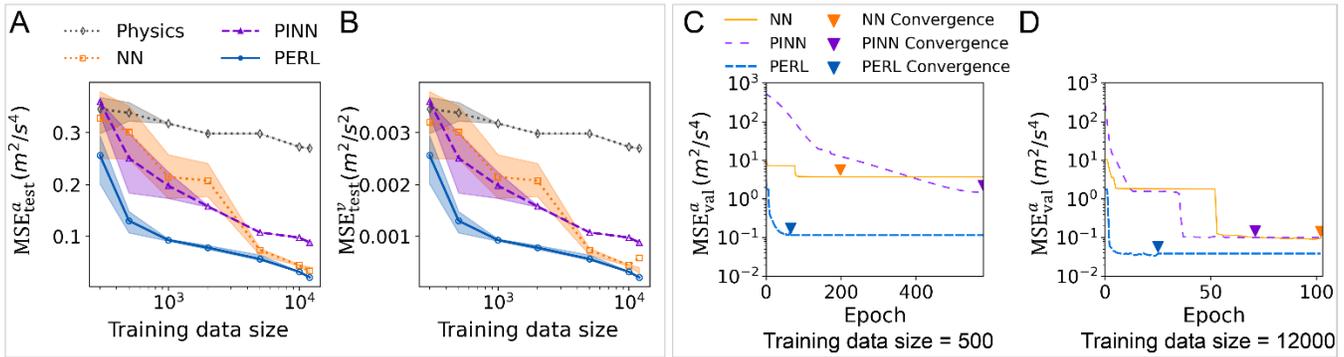

**Figure 8 PERL (LSTM+IDM) and baseline prediction performance.**

The performance of PERL (LSTM+FVD) is less effective than PERL (LSTM+IDM) with small amounts of data (Figure 9A 10B). This mainly results from the poor performance of the FVD model as a physical model when calibrated using small amounts of data. However, even though the physical model is not predictive, PERL (LSTM+FVD) still has a comparative prediction with the NN model and is better than the PINN model. Results also show the PERL model exhibits quick convergence rates for both small and large training data sizes (Figure 5C 5D).

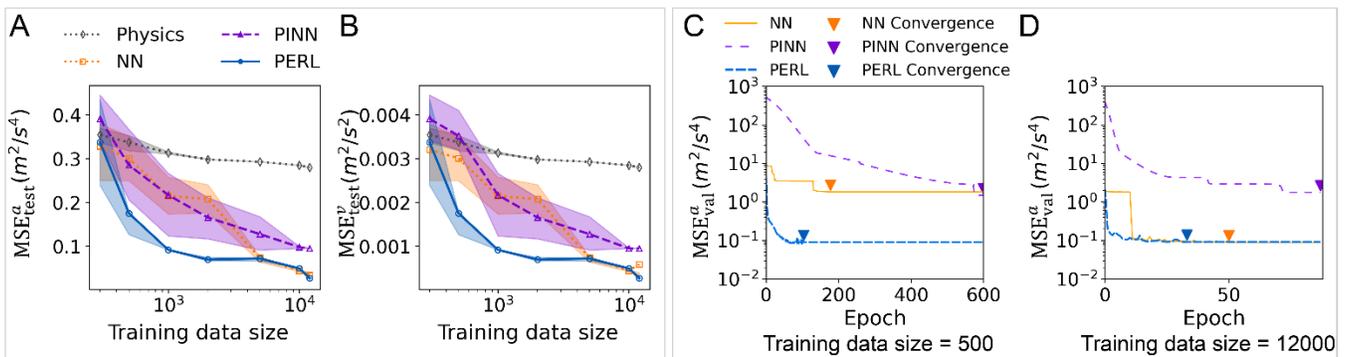

**Figure 9 PERL (LSTM+FVD) and baseline prediction performance.**

*5.4.2   Neural network architecture*

To validate the versatility of the PERL model for other deep learning models, we replaced the LSTM component of PIDL with a Gated Recurrent Unit (GRU) model, which is apt for time series prediction due to its capability to capture long-term dependencies in sequential data.

Contrasting the GRU-based model with the LSTM-based model revealed a notable difference. The GRU-based NN model lagged behind the LSTM-based NN model (Figure 10A 11B 11C 11D). The NN model (GRU) only surpassed the physics model (Adapted Newell) in acceleration prediction performance when the training data size reached 10,000. This disparity can be attributed to the inherent simplicity of the GRU structure compared to LSTM. GRU is prone to challenges like vanishing or exploding gradients, especially when processing extended sequences. Such a structure might inadequately capture intricate sequence patterns or long-term dependencies, leading to its inferior performance relative to LSTM. Besides, the GRU-based PERL (GRU+Adapted Newell) also trailed behind the LSTM-based PERL (LSTM+Adapted Newell) in acceleration prediction performance.

Despite the inherent simplicity of the GRU model, which resulted in diminished efficacy across GRU-based PERL, NN model, and PINN models, the PERL (GRU+Adapted Newell) still achieved a smaller prediction error than both the NN model and PINN with a limited dataset (fewer than 500 training data) (Figure 10A 11B 11C 11D). With sufficient data, both one-step and multi-step predictions by PERL yielded minimal prediction errors, comparable to those achieved by NN and PINN models.

The GRU-based PERL converges faster than baselines in various sample sizes (Figure 10E 11F 11G 11H). This reaffirms the earlier conclusion that acceleration residual learning converges faster than direct acceleration learning—and applies to both LSTM and GRU models.

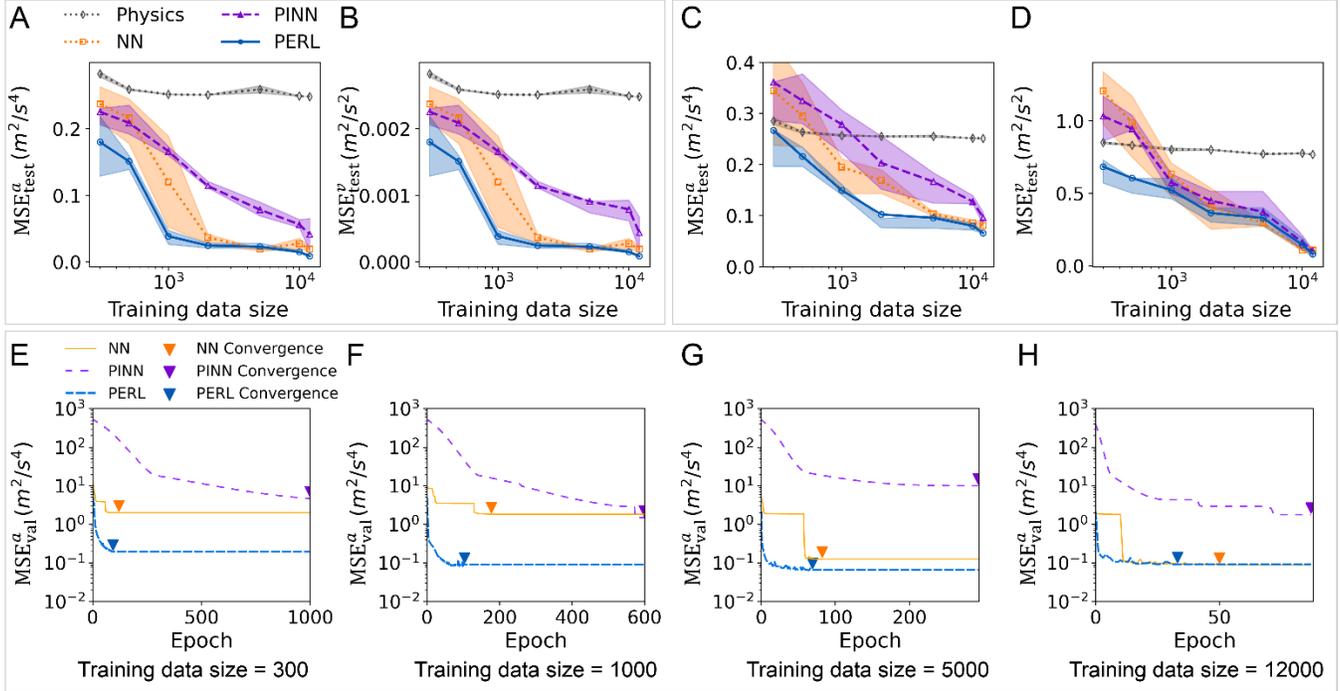

**Figure 10 PERL (GRU+Adapted Newell) and baseline prediction performance.**

## 6 CONCLUSION AND FUTURE WORKS

This study presented the novel PERL model as a potent tool for trajectory prediction, aiming to address inherent limitations found in existing physics models and NN models. The PERL model leverages the power of residual learning, a concept relatively unexplored in the traffic domain, providing the research community with a new perspective on traffic modeling and prediction.

We compared the predictive performance of the PERL model with traditional physics and NN models, using both HV and AV trajectory data. The results demonstrated the superiority of the PERL model in both one-step and multi-step acceleration prediction tasks. Notably, the PERL model consistently outperformed all other tested models across different data sources and vehicle types. PERL also has faster initial convergence during the training process than the traditional NN model and PINN. In sensitivity analysis, we validate the comparable performance of PERL using other residual learning models and physics car-following models.

Future research could focus on further exploring the advantages of the PERL model structure and addressing potential limitations to enhance its practical applicability. First, the robustness of PERL against biased physics models presents an exciting area for exploration. This characteristic has not been thoroughly investigated in this study, but it presents a promising avenue for future research. Systematic evaluations of the performance under different bias levels can provide valuable insights, along with developing techniques to mitigate or correct bias in the physics model. Second, it is important to recognize a potential constraint of the PERL model: its robustness against noisy data. Future work would conduct comprehensive investigations to assess its performance under noisy data conditions. Moreover, we will explore and develop novel techniques and methodologies to enhance its ability to effectively handle and mitigate the adverse effects of noisy data.

# 7 DECLARATION OF INTERESTS

The authors declare that they have no known competing financial interests or personal relationships that could have appeared to influence the work reported in this paper.

# 8 ACKNOWLEDGMENTS

This work was supported by the National Science Foundation Cyber-Physical Systems (CPS) program. Award Number: 2313578.